\documentclass[letterpaper, 10 pt, conference]{ieeeconf}  

\IEEEoverridecommandlockouts                              

\usepackage{graphicx} 
\usepackage{amsmath} 
\usepackage{amssymb}  
\usepackage{booktabs}
\usepackage[table,xcdraw]{xcolor}
\usepackage{tikz}
\usepackage{grffile}

\title{\LARGE \bf
	N-QGN: Navigation Map from a Monocular Camera using Quadtree Generating Networks
}

\author{Daniel Braun$^{1}$, Olivier Morel$^{1}$, Pascal Vasseur$^{2}$ and C\'edric Demonceaux$^{1}$%
	\thanks{$^{1}$Daniel Braun, Olivier Morel and C\'edric Demonceaux are with ImViA Laboratory, University of Bourgogne Franche-Comt\'e, 71200 Le Creusot, France.
		There email addresses are, respectively, {\tt\small daniel\_braun@etu.u-bourgogne.fr}, {\tt\small olivier.morel@u-bourgogne.fr} and {\tt\small cedric.demonceaux@u-bourgogne.fr}}%
	\thanks{$^{2}$Pascal Vasseur is with MIS Laboratory, University of Picardie Jules Verne, 80000 Amiens, France {\tt\small pascal.vasseur@u-picardie.fr}}%
}

\begin{document}

	\maketitle
	\thispagestyle{empty}
	\pagestyle{empty}

	\begin{abstract}
		
		Monocular depth estimation has been a popular area of research for several years, especially since self-supervised networks have shown increasingly good results in bridging the gap with supervised and stereo methods. However, these approaches focus their interest on dense 3D reconstruction and sometimes on tiny details that are superfluous for autonomous navigation. In this paper, we propose to address this issue by estimating the navigation map under a quadtree representation. The objective is to create an adaptive depth map prediction that only extract details that are essential for the obstacle avoidance. Other 3D space which leaves large room for navigation will be provided with approximate distance. Experiment on KITTI dataset shows that our method can significantly reduce the number of output information without major loss of accuracy.

	\end{abstract}

	\section{INTRODUCTION}
	
	Depth estimation represents a fundamental task for scene understanding in autonomous navigation. Self-supervised monocular depth networks \cite{godard2019digging} are proposing to estimate the depth based on a single image. It is geometrically an ill-posed problem since the depth information must be obtained from images by triangulating points observed from diverse location through displacement or by using a stereo-camera system. Yet, deep learning has proven its capability to efficiently predict it from a single image if it has been trained long enough on similar situations. In the same way as humans are capable of apprehending objects location by watching a video or by looking at an image.  
	
	When building a navigation map, the objective is to be able to avoid obstacles, and the knowledge of tiny details is not needed. This is why a dense depth map is sometimes an excessive solution because it will produce a too large point cloud to be computed in real time by an embedded system. Generally used solutions are to work with sparse information, to reduce the image resolution or to only extract close range 3d points. The main point is that the same degree of resolution at close distance and at long distance is not required and must be adapted. This is where quadtree structures can come to be of great benefit. Quadtree is a hierarchical way of representing 2d space inside a tree data structure. Each node of the tree represents a square area in the image with a location, a size and a value. Each node can be subdivided into four child nodes which will in turn describe a sub-area of the parent node. This type of representation allows to only explore into details some areas of the 2D space and to store coarse informations in areas where details are not required.
	
	\begin{figure}[t]
		\vspace{5px}
		\centering
		\includegraphics[width=0.48\textwidth]{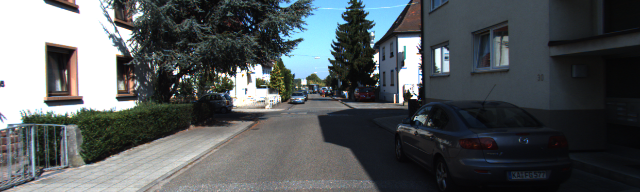}
		
		\includegraphics[width=0.48\textwidth]{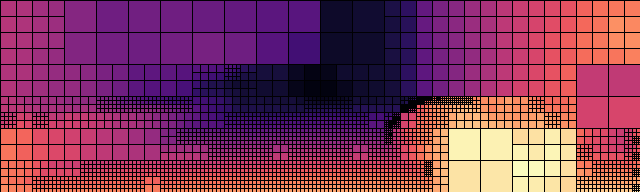}
		
		\includegraphics[width=0.48\textwidth]{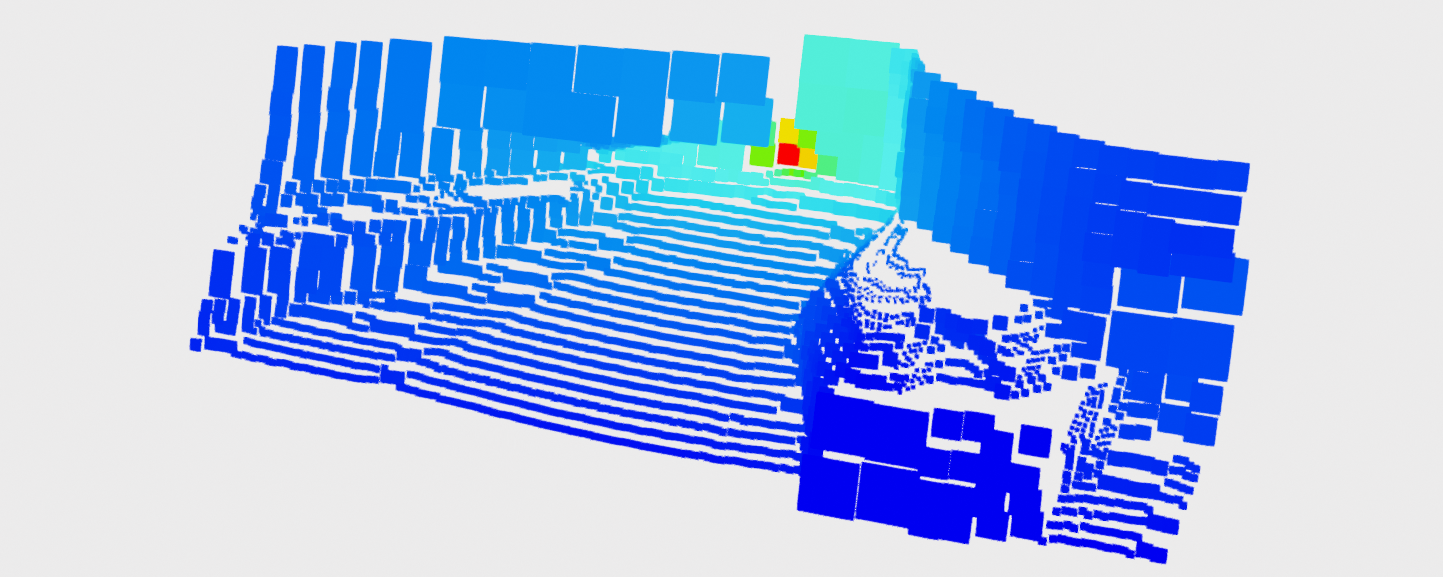}
		
		\label{vitrine}
		\caption{From top to bottom, the input image, the output depth map quadtree and the 3D projection of the quadtree depth informations.}
	\end{figure}
	
	In this paper, we propose a depth map quadtree estimation by adapting the Quadtree Generative Network (QGN) by Chatta et al. \cite{chitta2020quadtree} . QGN is based on Submanifold SparseConvNet \cite{3DSemanticSegmentationWithSubmanifoldSparseConvNet} and take advantage of the sparse convolution to output a quadtree from the network decoder. It is initially implemented for image segmentation and demonstrated its capability to reduce the computational need by conserving the prediction quality. Our approach is novel in the way that we are not working with classes, where the quadtree subdivision can be relatively obvious, but with continuous depth value. Accordingly, a criterion will be introduced to rule the manner the quadtree will be subdivided. For the training, we will use the state-of-the-art loss function for monocular depth prediction with the addition of this subdivision criterion.
	
	In the rest of the paper, we will often refer to the classical depth map prediction methods as being a \textit{dense} depth map prediction as opposed to quadtree prediction. The term \textit{dense} is used here because it is a prediction for each and every pixel in the image, whereas the quadtree offers a compressed representation where a single value can describe a large area in the image, although the final prediction is still dense.

	\section{RELATED WORKS}
	
	\subsection{Monocular depth estimation}
	
	Monocular depth map prediction is an inverse problem whose goal is to find the solution to a problem where half of the information normally needed is missing. From a single RGB image, we predict a depth map describing the scene that allows us to find the missing image pair using photometric reprojection. The capability of such method was first demonstrated by Eigen et al. \cite{eigen2014depth} and shortly improved it a year later \cite{Eigen_2015_ICCV} by taking full advantage of the multi-scale prediction. The method has been progressively improved the following years with the work of Liu et al. \cite{liu2015deep} or Roy and Todorovic \cite{roy2016monocular}. But the field of study truly emerged with the arrival of a self-supervised method proposed by Garg et al. \cite{garg2016unsupervised}. Indeed, training a network in a self-supervised way makes it possible to do without annotated data and thus greatly extends the possibilities. There is no more limitation to the few annotated datasets where ground truth can sometime be incomplete or erroneous due to human error during annotation. 
	Godard et al. \cite{godard2017unsupervised} pushed the method a step further by introducing the left-right consistency in the loss. If you know the depth and one of the image, you can recreate the adjacent view by warping and minimize the photometric error between the projection and the actual image. Zhou et al. \cite{Zhou_2017_CVPR} proposed to use the ego-motion for training and so open the gate to depth learning with a single camera. It works by having two network simultaneously learning the depth map and the relative pose between two images at adjacent time frame. In 2019, Godard proposed Monodepth2 \cite{godard2019digging} which combined the previous work \cite{godard2017unsupervised} including the novelty proposed by Zhou et al. with the addition of some ways in order to better handle occlusion. The method is two years old now but stays one of the most efficient self-supervised method is this field, even if some others  managed to outperform it \cite{yang2020d3vo, peng2021excavating}.
	
	\subsection{Quadtree and octree representation}
	
	Quadtree is a hierachical tree data structure that is deeply tight to the image processing field since it was first introduced nearly 40 years ago by Samet et al. \cite{samet1984quadtree}. It is a convenient way to efficiently compress data because a single node of the tree can describe a large area in the image. Therefore, this method has often been used for image compression especially for navigation applications \cite{wang2018quadtree}. It is only recently that quadtrees were employed in deep learning \cite{jayaraman2018quadtree, chitta2020quadtree} probably because its unregular data structure makes it hard to fit the regular data tensor representation. In fact, the quadtree deep learning approaches are derived from the octree methods \cite{tatarchenko2017octree, liu2020rocnet} that came on earlier. Octree data structures are the equivalent in 3D of the quadtree and are largely used in 3D representation. Indeed, the gain in compression is even larger because a 3D environment is mainly composed of empty volume, which allows to construct large global maps \cite{hornung2013octomap}.

	\section{METHOD}
	
	In this section, we discuss our network to predict an efficient quadtree navigation map. We present the quadtree representation, how it can be directly inferred using the Quadtree Generative Network \cite{chitta2020quadtree} and how to adapt it to a depth prediction approach.
	
	\begin{figure}[h]
		\centering
		\includegraphics[width=0.48\textwidth]{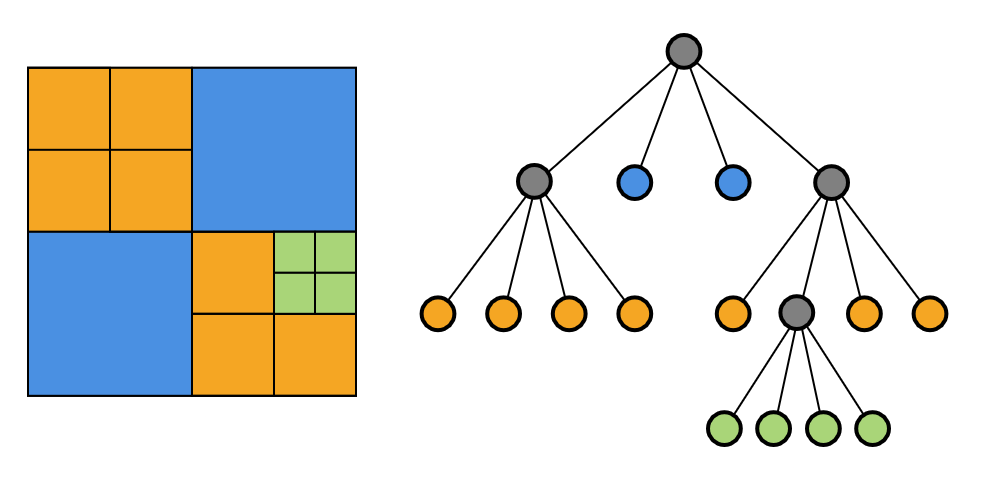}
		
		\caption{On the left side is the quadtree image. Its corresponding hierarchical tree representation is on the right. Each node store a single value but represent a different area on the image depending on its location in the tree. Gray nodes represent inner nodes that have been subdivided.}
		\label{fig:quadtree}
	\end{figure}
	
	\subsection{Quadtree and tree-pyramid representation}
	\subsubsection{Quadtree representation}
	
	A quadtree is a hierarchical tree data structure composed of nodes as illustrated in Figure \ref{fig:quadtree}. Each node is representing an square area of the image and can be subdivided into four children nodes each representing one of the sub-areas. Each node is designated by its depth on the tree, its location on the image and its stored value corresponding to the region of pixels it describe. It is noted $q_i = \{l_i, x_i, y_i, v_i\}$ where $\{l_i, x_i, y_i\}$ is the level and spatial location and  $v_i$ the corresponding value. The complete quadtree is defined by
	
	\begin{equation}
		Q = \{q_1, q_2, ..., q_n\} \ \text{with} \ n \in \mathbb{N}
	\end{equation}
	
	The advantage of such a structure is that a single element in the quadtree can represent a large area of the image. In that way, we can highly reduce the number of stored data and obtained a compressed representation of our scene. This compression is smarter than a simple down sampling because it allow us to really focus our attention on the important part of the image.

	\subsubsection{tree-pyramid representation}
	\label{sssec:tree_pyramid}
	A tree-pyramid is a particular case of quadtree where the tree is complete. It means every possible node of the tree contains value. In the case of a tree-pyramid representing an image, it corresponds to have a value affected to each pixel. That's the heaviest a quadtree can be but has the advantage of making two quadtrees simple to compare, because we don't have to account for the particular branching of each tree. During training, we will work with the tree-pyramid representation to evaluate our prediction using dense approaches, as if our network predicted a value per pixel.
	
	\begin{figure}[h]
		\vspace{5px}
		\centering{
			\includegraphics[width=0.48\textwidth]{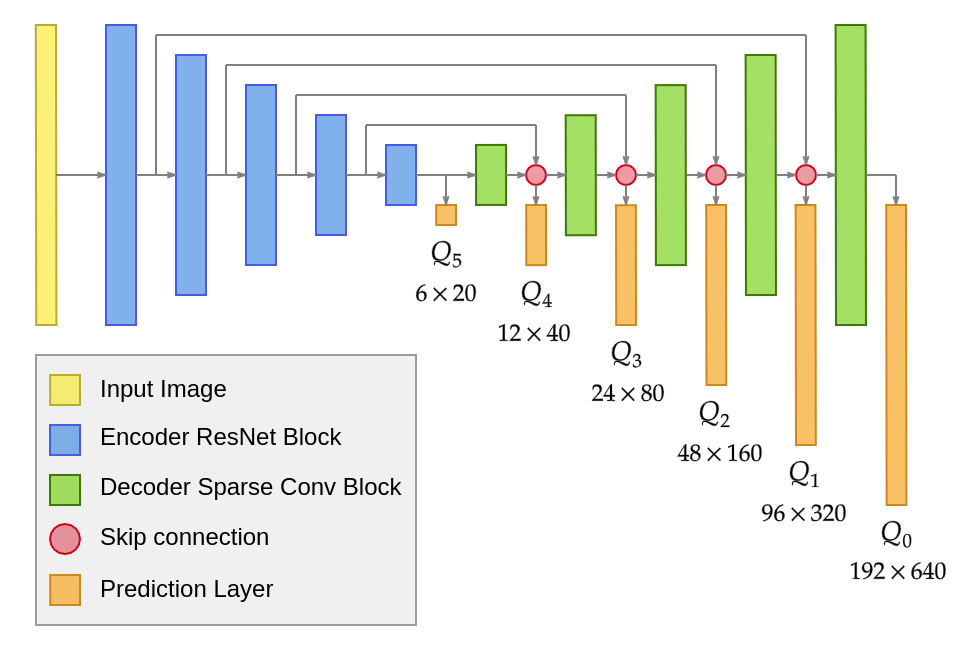}
			\caption{The network has a U-Net architecture composed of a dense ResNet 18 encoder in blue and a Submanifold sparse decoder in green. The prediction layers are in orange below the decoder. Each one of them outputs a depth prediction of the scene at various resolutions, from $Q_5$ to $Q_0$. $Q_5$ represents the root of the tree and $Q_0$ the leaf nodes. All together they are forming the quadtree representation.}
			\label{fig:network}
		}
	\end{figure}
	
	\subsection{The network}
	The network is an end-to-end solution that converts an RGB image into a quadtree representation of a depth map for navigation purpose. The architecture is presented on Figure \ref{fig:network} and is based on the Quadtree Generative Network (QGN) proposed by Chitta et al. \cite{chitta2020quadtree} and the Monodepth2 architecture from Godard et al. \cite{godard2019digging}.
	
	\subsubsection{Architecture}
	
	In the way that we are trying to infer depth information from a single image, our approach is similar to the state-of-the-art concerning the depth map prediction. This is why we decided to use the same type of ResNet encoder that is generally used. The decoder is derived from QGN sparse decoder and adapted to fit our goal in a sense that we are not in a multiclass segmentation approach but in depth prediction so only a single channel output is considered.

	\subsubsection{Sparse Convolution}
	As presented in \cite{3DSemanticSegmentationWithSubmanifoldSparseConvNet}, Submanifold sparse convolutional approaches are meant to efficiently operate on spatially-sparse data by ignoring the empty area during the convolutional process. Therefore it only operates on active sites which allows to greatly reduce the computational and memory requirements. A site is either considered inactive because no data is available or because it has been willingly threshold out during the process.
	
	In our case, we have dense information at the entry of the decoder, that we will voluntarily sparsify after each layer of the decoder. It is achieved by applying a mask on the features to eliminate superfluous data. This mask is either induced based on the predicted depth  or can be predicted by the network. For the case of image segmentation as in \cite{chitta2020quadtree}, it can be done in a straight forward way by having a mixed class in the prediction. If the mixed class obtains the highest score, it means the area is still composed of multiple classes and will be subdivided. However, in the case of a depth information prediction, we are dealing with continuous values. This subdivision should classically be performed by knowing the dense informations. But the whole idea is to avoid having to compute dense data and directly infer the quadtree. 
	
	\begin{figure*}[h]
		\vspace{5px}
		\centering
		\includegraphics[width=\textwidth]{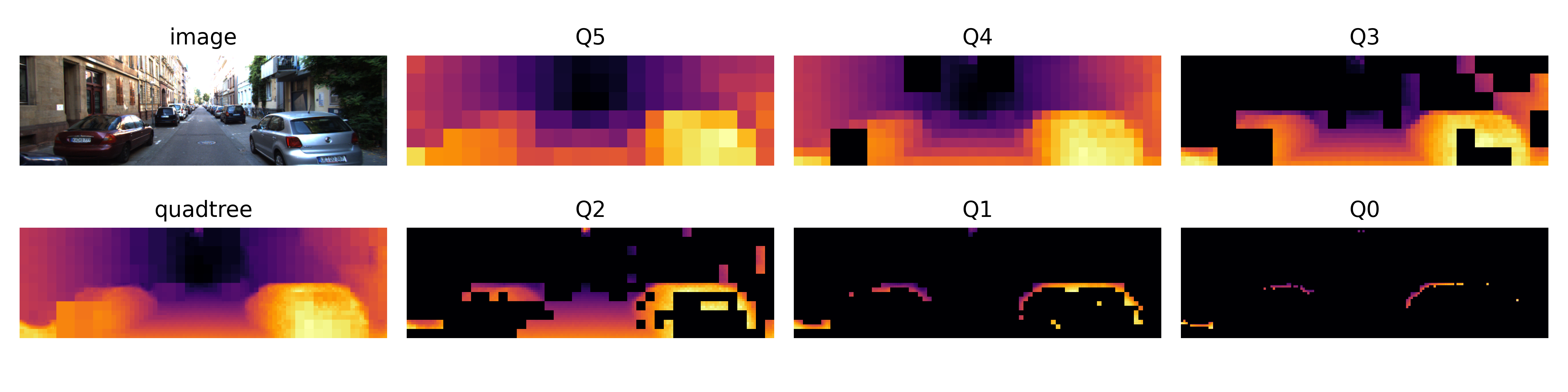}
		\caption{Quadtree decomposition. On the far left is the input image at the top and quadtree at the bottom. $Q_5$ to $Q_0$ are the outputs of the network and represent each level of the quadtree. The black areas in the images represent the inactive sites, i.e. the zones where more details were not necessary according to the subdivision criteria defined in the Equation \ref{eq:sub_crit}.}
		\label{fig:quadtree_decomposition}
		
	\end{figure*}
	
	\subsection{Quadtree decomposition}
	The quadtree is constructed, through the decoder as presented in Figure \ref{fig:network}, from the root of the tree (low resolution) to the deepest leaves (high resolution). Each level of the quadtree can be represented as an image noted $Q_5$ for the lowest resolution to $Q_0$ for the highest resolution as represented in Figure \ref{fig:quadtree_decomposition}. Since the features are more and more sparse though out the decoder, $Q_0$ will have a very low density and mainly describes edges. On the contrary, $Q_5$ will be a dense low resolution prediction describing the general shape of the map.
	Since the quadtree is constructed from low resolution to high resolution, the decision to split a node in the tree must be made without the knowledge of all the children values, which is in contradiction with how the quadtree construction generally works. It makes it an ill-pose problem where we have to make decisions based on incomplete data. To address this issue, few solutions can be applied. We could work around the problem estimating the four direct children of each node and decide if the node subdivision is necessary from there or make the network learning how to subdivide the quadtree with a ground truth supervision.
	
	\subsubsection{Multi-scale prediction}
	The multi-scale depth map prediction method is particularly well fitted for quadtree prediction because it is in essence a multiscale representation. During training, the multi-scale photometric error minimisation loss from self-supervised depth prediction can be directly used since it is possible to extract multi-scale representation from a quadtree. 
	
	As illustrated in Figure \ref{fig:quadtree_decomposition}, the complete quadtree can be represented as a combination of multi-resolution images. Except than instead of recomputing everything at each scale, only the parts of the images where details are needed is refined. It means we can simply access at lower resolution quadtree, noted $Q_{5 \rightarrow n}$, which combine the informations from $Q_5$ to $Q_n$ with $n \in \{4,3,2,1,0\}$. This operation is defined by:
	
	\begin{equation}
		Q_{5 \rightarrow n} = Q_n + \sum_{i=n+1}^5 Q_i \cdot (1 - A_{i-1})
		\label{eq:complete_quadtree}
	\end{equation}
	where $A_{i-1}$ is the active sites map of $Q_{i-1}$ with values in $A_{i-1} \in \{0, 1\}$. Concretely speaking, we keep values from a specific level of the quadtree $Q_i$ only on the inactive sites location of its direct lower level $Q_{i-1}$ and we keep every informations from the lowest level desired $Q_n$.
	
	\subsubsection{Photometric minimization}
	Multi-resolution quadtree under a tree-pyramid representation (see \ref{sssec:tree_pyramid}) can be considered as a full pixel-wise depth map. In that sense, we can apply the same loss function from \cite{godard2019digging}, which is a combination of a photometric error loss $\mathcal{L}_p$ and a per-pixel smoothness loss $\mathcal{L}_s$:
	
	\begin{equation}
		\mathcal{L} = \frac{1}{N} \sum_i^N (\mu \mathcal{L}_p^i + \lambda \mathcal{L}_s^i)
	\end{equation}
	with $N$ being the number of levels in the quadtree, here 6 in our approach. $\mu$ and $\lambda$ are constants value chosen empirically.
	
	The photometric reprojection error, noted $pe$, between a source $I_t$ and a target image $I_{t'}$ can be expressed as
	
	\begin{equation}
		\mathcal{L}_p = \min_{t'}{pe(I_t, I_{t' \rightarrow t})}
	\end{equation}
	with $I_{t' \rightarrow t}$ being the projection of $I_{t'}$ into the frame $t$ based on the knowledge of the predicted depth, the relative pose and the camera intrinsic parameters. The photometric error is a combination of L1 norm and SSIM such as 
	
	\begin{equation}
		pe(I_a, I_b) = \frac{\alpha}{2} (1 - \text{SSIM}(I_a, I_b)) + (1 - \alpha) \|I_a - I_b\|
	\end{equation} 
	where $\alpha$ is a constant value set to $0.85$.
	
	The smoothness loss in equation \ref{eq:smoothness_loss} term is sharpening the edges and is encouraging smooth values on texturless areas. As a result, the depth prediction will have smaller variations along the value which will lead the quadtree to be smaller because less subdivision will be required to describe the scene (see \ref{ssec:quadtree_subdivision}).
	
	\begin{equation}
		\mathcal{L}_s = | \partial_x d_t^* | e^{|\partial_x I_t|}  + |\partial_y d_t^*| e^{-|\partial_y I_t|},
		\label{eq:smoothness_loss}
	\end{equation}
	with $d^*_t = d_t / \bar{d_t}$ being the mean-normalized inverse depth to prevent the depth to diminish \cite{wang2018occlusion}.

	\subsubsection{Subdivision criterion}
	\label{ssec:quadtree_subdivision}
	Depth prediction for navigation aims at accurately detect obstacles at short range because it needs to have the knowledge of any imminent obstacle. The long range prediction helps to increase the map accuracy in the long time by providing a first glance at what is expected, but is generally rightly considered with a high uncertainty coefficient. Subsequently we choose a quadtree subdivision criterion which favor short-range accuracy by being based on the disparity value. The disparity is equivalent to the inverse of the depth, so a small variation in the disparity value will have more impact at long range than at short range.

	The quadtree must be constructed without the complete knowledge of the dense information. It is possible to only consider the four direct children of each node  and decide based on those values if it's beneficial for the node to be subdivided. The node will be split if the highest difference between the four children values ($v_i, i=1..4$) exceed a threshold $\tau$. On the image, it will consist of applying patches $\mathcal{A}$ of size 2x2 with a stride of 2,

	\begin{equation}
		\mathcal{A} =
		\begin{cases}
			\mathbf{1}_{2\times2} & \text{if} \max(v_i) - \min(v_i) > \tau, \\
			\mathbf{0}_{2\times2} & \text{else}.
		\end{cases}
		\label{eq:sub_crit}
	\end{equation}

	Depending on the threshold result, the node will be subdivided or not. Therefore the results will define the activation map of the following layer in the decoder. However, it will have no impact on the already predicted values regardless of the criterion results. 
	
	Since the children have been computed, we can choose to keep the values anyway and stop the subdivision afterward or to discard those values to only keep the value from the father node. It mostly depend on what we are trying to achieve. We can either have the smallest possible quadtree that fulfill our criterion or we accept to have a bit larger quadtree with an increase of accuracy. When working with quadtrees, it is always a trade-off between size and precision. We choose the second solution to keep all the predicted values and adjust the subdivision criterion to lighten the quadtree.

	\subsubsection{Scale invariance}
	The subdivision criterion is based on the inverse depth value, which means the scale has a major impact on the subdivision decision. Yet, monocular depth prediction is an up-to-scale prediction even if the scale is learned during the training. In fact, a variation in the scale between the stereo training, the monocular training or the combination of the two. In monocular training, scale variation can occur depending on how the couple depth and pose have been optimized together. In our application, a stable scale is required, since it is linked to the subdivision criterion. For this reason, the training has to be performed with a calibrated stereo pair.

	\section{EXPERIMENTS}
	
	In this section, we will discuss about the capability of our end-to-end network called N-QGN (Navigation-Quadtree Generating Network) to efficiently construct a quadtree navigation map and how it competes in respect with a quadtree constructed from dense information.
	
\begin{figure*}[t]
		\vspace{5px}
		\resizebox{\textwidth}{!}{%
			\begin{tabular}{cccccc}
				\includegraphics{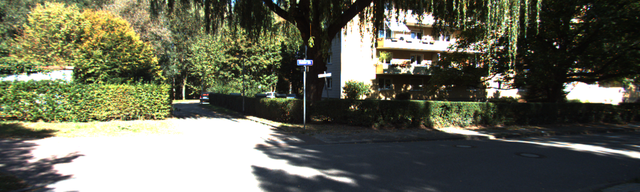} &
				\includegraphics{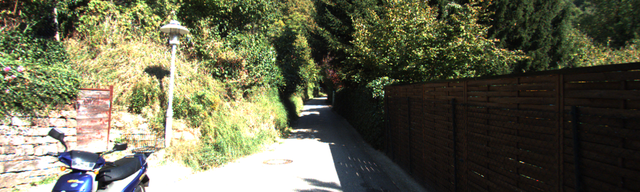} &
				\includegraphics{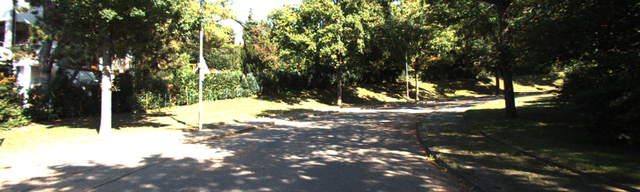} &
				\includegraphics{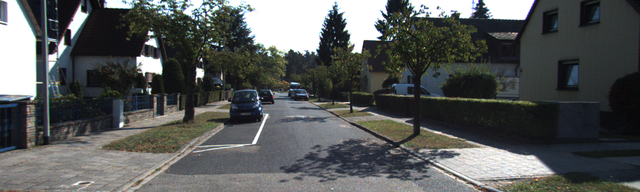} \\ \\
				\includegraphics{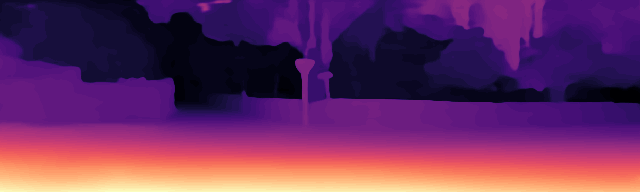} & 
				\includegraphics{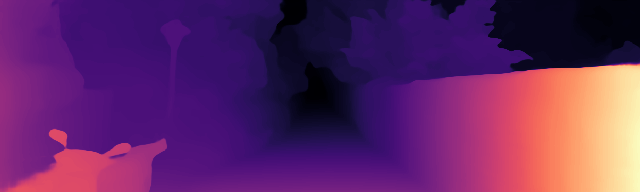} &
				\includegraphics{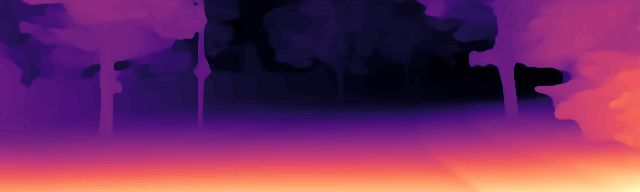} &
				\includegraphics{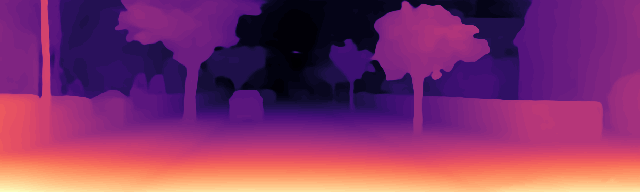} \\ \\
				\includegraphics{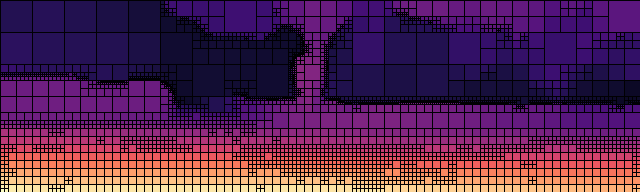} &
				\includegraphics{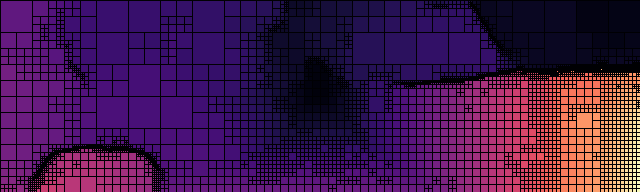} &
				\includegraphics{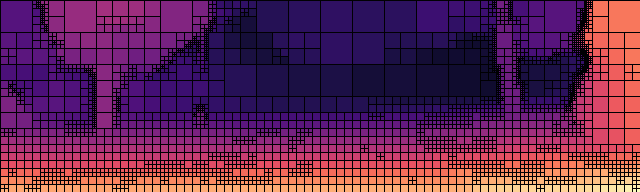} &
				\includegraphics{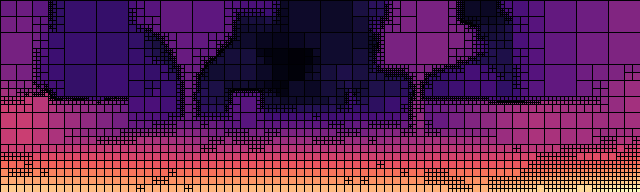} \\ \\
				\includegraphics{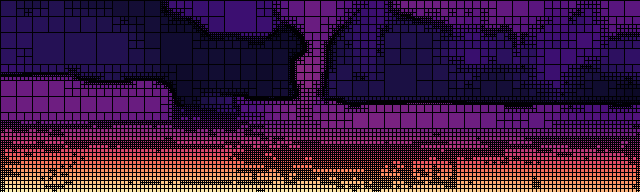} &
				\includegraphics{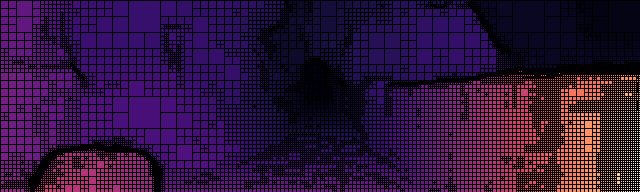} &
				\includegraphics{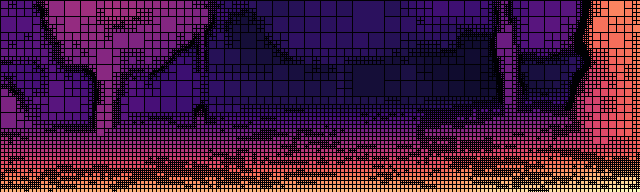} &
				\includegraphics{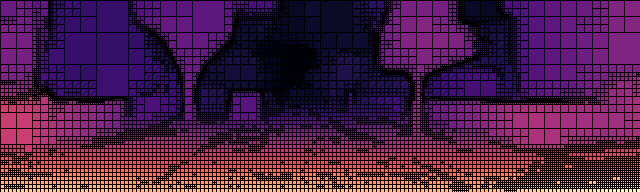} \\ \\
				\includegraphics{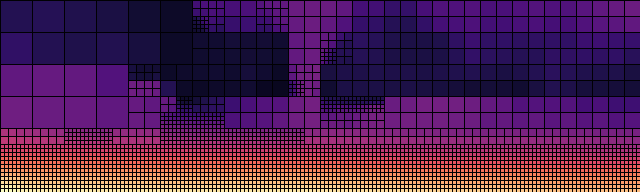} &
				\includegraphics{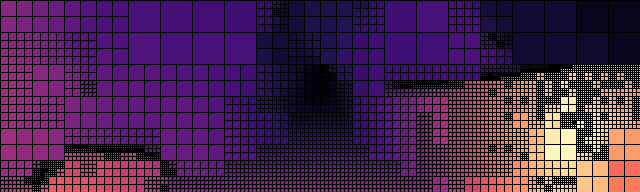} &
				\includegraphics{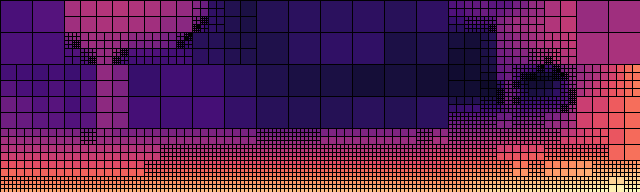} &
				\includegraphics{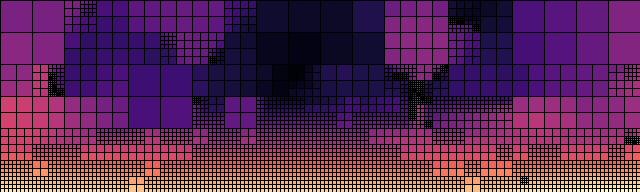} \\ \\
				\includegraphics{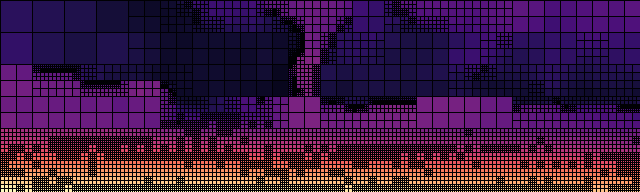} &
				\includegraphics{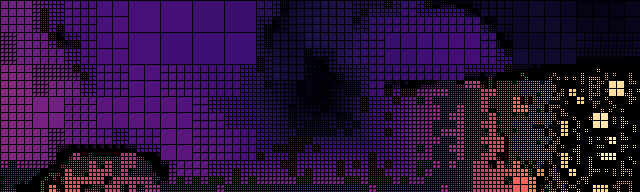} &
				\includegraphics{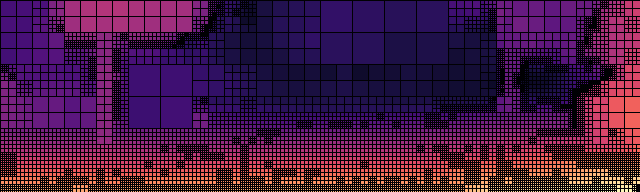} &
				\includegraphics{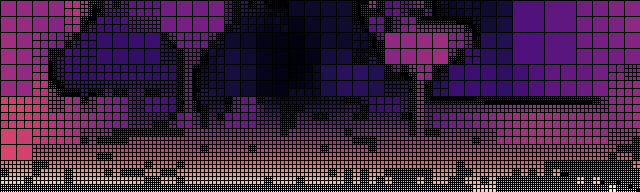} \\ 
				
			\end{tabular}
		}
		\caption{Illustration of the results on four indenpendant images from the testing set. From top to bottom: the input image, the stereo reference map \cite{cheng2020hierarchical}, the next two lines are Monodepth2 \cite{godard2019digging} converted into quadtree with a compression rate of respectively 33.0 and 10.7 and the last two lines are our method N-QGN with a compression rate of respectively 30.9 and 11.6. The compression rate is averaged over a set of validation images and is dependent of the geometry of the scene.}
		\label{fig:global_results}
	\end{figure*}

	Experiments were conducted on the Kitti dataset \cite{geiger2012we} since it is one of the most commonly used dataset for navigation on depth map prediction where stereo images can be used for fair comparisons with geometric methods. Results on each evaluated methods are presented in Figure \ref{fig:global_results} on samples images from the testing set.

	\subsection{Quadtree Navigation map}
	As presented before, the objective of our network is to produce a light and accurate navigation map without extra superfluous details. What is meant is we know that predicting a depth information from a single image will not be as accurate as from a stereo pair or from a sequence of images through motion. And the further the obstacle is, the less accurate our prediction will be. The whole interest is to have an end-to-end method that infers a quadtree representation of a depth map directly from a RGB image.

	To the best of our knowledge, our method is the only one to predict depth map information under a quadtree data structure with single image. Thus, it is not possible to compare directly with the state of the art. Therefore, the comparison will be done with the dense depth map methods from mono-camera called Monodepth2 \cite{godard2019digging}, from which the training procedure is based. In order to have a common evaluation ground, they will be converted into quadtrees in order to match with our method.
	The reference map is constructed from one of the most advanced method for stereo disparity estimation, namely LEAStereo from Cheng et al. \cite{cheng2020hierarchical}.

	\begin{table}[t]
		\vspace{5px}
		\centering
		\caption{Quadtree structure likelihood. }
		\label{tab:quadtree_likelihood}
		\begin{tabular}{ccccccccc}
			\toprule
			& Compression & quadtree structure \\
			Method & ratio & likelihood$\uparrow$ \\ \midrule
			monodepth2\cite{godard2019digging} + quadtree &  33.0  &  \textbf{95.2\%} \\  
			N-QGN &   30.9  & 94.2\% \\ \midrule
			monodepth2\cite{godard2019digging} + quadtree & 10.7 & \textbf{82.2\%} \\
			N-QGN & 10.2 & 81.3\% \\
			\bottomrule
		\end{tabular}%
	\end{table}

	\subsubsection{Quadtree structure likelihood}
	The quadtree data structure is evaluated by comparing the node distribution between our prediction and the quadtree constructed from the reference map. The results are presented in table \ref{tab:quadtree_likelihood} and highlight the capability of our network to correctly construct a quadtree even from incomplete data. Indeed, the results are of the same order as the dense prediction converted to quadtree, yet with a slightly lower value.

	\begin{table}[]
	\vspace{5px}
	\centering
	
	\caption{Evaluation of the methods against a reference map on the Kitti 2012 dataset. The methods are compared with two different compression ratios. Our results are confronted to Monodepth2 converted into a quadtree.}
	\label{tab:quadtree_comparison}
	\resizebox{0.485\textwidth}{!}{%
		\begin{tabular}{ccccccccc}
			\toprule
			& compression & \\
			Method & ratio   & Abs Rel$\downarrow$ & Sq Rel$\downarrow$ & RMSE$\downarrow$ \\ \midrule
			monodepth2\cite{godard2019digging} + quadtree &  32.2  &     \textbf{0.155}  &   2.284  &  11.071  \\ 
			N-QGN &   30.9  &   0.163  &   \textbf{2.106}  &   \textbf{9.737} \\ \midrule
			monodepth2\cite{godard2019digging} + quadtree &  10.7   &   \textbf{0.152}  &   2.246  &  10.990  \\
			N-QGN &  10.2 &    0.157  &   \textbf{2.062}  &  \textbf{9.765}  \\
			\bottomrule
		\end{tabular}%
	}
\end{table}

	\subsubsection{Depth evaluation}
	
	For the depth prediction evaluation, we compare our quadtree to the dense reference 
	by using classical metrics, such as the absolute relative error, the square relative error and the root mean square error (RMSE).
	The results are available in the table \ref{tab:quadtree_comparison} and compared with Monodepth2+quadtree  on two different compression ratios. This compression ratio is related to the parameter $\tau$, introduced in equation \ref{eq:sub_crit}, and defined by:
	\begin{equation}
		\text{Compression ratio} = \frac{\sharp (D)}{\sharp(Q)}
		\label{eq:comp_rate}
	\end{equation}
	with $\sharp(D)$ is the number of pixels in the dense depth map (image resolution) and $\sharp(Q)$ is the number of nodes of the quadtree depth map.
	This number  represents the ratio between the number of elements in a dense prediction (uncompressed) over the number of elements in a quadtree prediction (compressed). A high compression rate means a smaller quadtree so a coarser representation which explains the higher error when confronted to the dense reference. We can note that the RMSE is high for every method because we are comparing our approach with a dense stereo depth reference where every pixels in the image are evaluated, including the sky which is known as badly reconstructed by self-supervised photometric based methods. In the end, our method demonstrates its capability to efficiently construct a quadtree depth representation.

	\begin{table}[]
		\vspace{5px}
		\centering
		
		\caption{Data distribution along the quadtree levels from the root nodes in $Q_5$ to the leaf nodes in $Q_0$ for our method N-QGN.}
		\label{tab:quadtree_density}
		\resizebox{0.485\textwidth}{!}{%
			\begin{tabular}{ccccccc}
				\toprule
				Compression & $Q_5$ & $Q_4$ & $Q_3$ & $Q_2$ & $Q_1$ & $Q_0$   \\ \midrule
				30.9 & 18.24\% & \textbf{32.15\%} & 22.01\% & 24.58\% & 2.42\% & 0.61\% \\
				10.2 & 7.56\%  &  23.03\%  &  21.58\%  &  \textbf{31.82\%}  &  13.07\%  &  2.94\% \\
				\bottomrule
			\end{tabular}%
		}
	\end{table}
	
	\begin{table}[t]
		\vspace{5px}
		\centering
		
		\caption{Comparison of model size and complexity}
		\label{tab:size_complexity}
		\begin{tabular}{cccccccc}
			\toprule
			Method & FLOPs & Parameters   \\ \midrule
			monodepth2 \cite{godard2019digging} &  8.0G & 14.842M \\ 
			N-QGN (comp. ratio = 30.9) & 5.3G & 13.115M  \\ 
			N-QGN (comp. ratio = 10.2) & 5.7G & 13.115M \\
			\bottomrule
		\end{tabular}%
	\end{table}
	
	\subsection{Memory footprint analysis}
	
	\subsubsection{Data distribution}
	The performances of the methods have been evaluated on previous sections at two data compression factor. We analyze here in more detail this ratio and its distribution along the different levels of the quadtree. It is done by observing the information distribution from $Q_5$ to $Q_0$ in the table \ref{tab:quadtree_density} over a set of validation images. The percentage associated with each $Q_i$ is the average percentage of information from that $Q_i$ that is found in the final quadtree. 
A high percentage for one of the $Q_i$ means this $Q_i$ largely describe the final quadtree.

	For the methods with the highest compression rate, more than 60\% of the information is predicted within the first three levels of the quadtree ($Q_5$, $Q_4$ and $Q_3$) and less than 1\% of the pixels reach the leaf nodes, i.e. the last layer of the decoder. For the method with the lowest compression rate, the information are centered around $Q_4$, $Q_3$ and $Q_2$. This solution will attach more attention to details and is probably less adapted to a light navigation solution than the previous solution.
	
	\subsubsection{Model size and complexity}
	Submanifold Sparse Convolutions \cite{3DSemanticSegmentationWithSubmanifoldSparseConvNet} present the advantage of only operate on active sites. Therefore, the complexity of the network is linked to the sparsity of the prediction. As illustrated in table \ref{tab:size_complexity}, the floating point operations (FLOPs) performed in the network and the number of parameters have been reduced compare to monodepth2 \cite{godard2019digging}. let us note that the cost to convert the monodepth2 into quadtree is not taken into account for the computation cost
	
	\section{CONCLUSIONS}
	
	In this paper, we presented an end-to-end solution to efficiently predict a quadtree representation of a depth map for navigation goal. This solution takes advantage of the Submanifold sparse convolution to eliminate superfluous information inside the network. Therefore, it is no longer necessary to use dense prediction to infer a quadtree. Sparse convolution was initially designed to efficiently process incomplete or discrete data. In that sense, the data can be willingly sparsify to produce sparse representation from dense data in order to focus the interest on relevant information. 
	
	The Monocular quadtree depth prediction is a double challenge as it combines the ill-posed problem of predicting the depth from a single image and the necessity to decide how to subdivide the quadtree nodes with a limited knowledge of the information. Yet, we came up with a simple and efficient solution to address this challenge. Deciding for the subdivision upon the knowledge of the four direct children of each nodes allows to reduce the problem complexity. As a result, we have a self-supervised method presenting good results.
		
	Our study focus on producing a light representation by deliberately choosing a strict subdivision criterion. 
	An alternative would be to predict the most accurate quadtree depth map to compete with the state of the art. Indeed, the last layers of the decoder only deal with a fraction of the image, which are mostly edges. So, it makes them more specialized to efficiently predict a specific information.

	\addtolength{\textheight}{-10.5cm}   
	



	\section*{ACKNOWLEDGMENT}
	This work is supported by the French National Research Agency ANR-18-CE33-0004-CLARA. We gratefully acknowledge the support of NVIDIA Corporation with the donation of GPUs used for this research.


	\bibliographystyle{IEEEtranS}
	\bibliography{ref.bib}

\end{document}